\documentclass[sigconf]{acmart}
\AtBeginDocument{%
  }

\copyrightyear{2026}
\acmYear{2026}
\setcopyright{cc}
\setcctype{by}
\acmConference[KDD 2026] {Proceedings of the 32nd ACM SIGKDD Conference on Knowledge Discovery and Data Mining V.2}{August 9--13, 2026}{Jeju Island, Republic of Korea.}
\acmBooktitle{Proceedings of the 32nd ACM SIGKDD Conference on Knowledge Discovery and Data Mining V.2 (KDD 2026), August 9--13, 2026, Jeju Island, Republic of Korea}
\acmISBN{979-8-4007-2259-2/2026/08} 
\acmDOI{10.1145/3770855.3817700}
\usepackage{siunitx}
\usepackage{multirow}
\usepackage{booktabs}
\usepackage{tabularx}
\usepackage{balance}
\usepackage{bibentry}
\usepackage{mathtools}
\usepackage{amsmath}
\usepackage{makecell}

\begin{document}

\title[AGMark: Attention-Guided Dynamic Watermarking for Large Vision Language Models]{AGMark: Attention-Guided Dynamic Watermarking\\for Large Vision Language Models}
\author{Yue Li}
\authornote{Both authors contributed equally to this research.}
\orcid{0009-0005-5509-2103}
\affiliation{%
  \institution{East China Normal University}
  \city{Shanghai}
  \country{China}
}
\email{yue\_li@stu.ecnu.edu.cn}

\author{Xin Yi}
\authornotemark[1]
\orcid{0009-0003-9646-1468}
\affiliation{%
  \institution{East China Normal University}
  \city{Shanghai}
  \country{China}
}
\email{xinyi@stu.ecnu.edu.cn}

\author{Dongsheng Shi}
\orcid{0009-0004-2059-5491}
\affiliation{%
  \institution{East China Normal University}
  \city{Shanghai}
  \country{China}
}
\email{51275901023@stu.ecnu.edu.cn}

\author{Yongyi Cui}
\orcid{0009-0006-5680-1570}
\affiliation{%
  \institution{East China Normal University}
  \city{Shanghai}
  \country{China}
}
\email{yycui@stu.ecnu.edu.cn}

\author{Gerard de Melo}
\orcid{0000-0002-2930-2059}
\affiliation{%
  \institution{Hasso Plattner Institute}
  \city{Potsdam}
  \country{Germany}
}
\affiliation{%
  \institution{University of Potsdam}
  \city{Potsdam}
  \country{Germany}
}
\email{gdm@demelo.org}

\author{Linlin Wang} 
\authornote{Corresponding author.} 
\orcid{0000-0003-0304-7560}
\affiliation{%
  \institution{East China Normal University} 
  \city{Shanghai} 
  \country{China} 
} 
\affiliation{%
  \institution{City University of Hong Kong} 
  \city{Hong Kong} 
  \country{China} 
}
\email{llwang@cs.ecnu.edu.cn} 

\begin{abstract}
Watermarking has emerged as a pivotal solution for content traceability and intellectual property protection in large vision language models (LVLMs). However, vision-agnostic watermarks may introduce visually irrelevant tokens and disrupt visual grounding by enforcing indiscriminate pseudo-random biases. Additionally, current vision-specific watermarks rely on a static, one-time estimation of vision-critical weights and ignore the weight distribution density when determining the proportion of protected tokens. This design fails to account for dynamic changes in visual dependence during generation and may introduce low-quality tokens in the long tail. 
To address these challenges, we propose  Attention-Guided Dynamic Watermarking (AGMark), a novel framework that embeds detectable signals while largely preserving visual-semantic fidelity. At each decoding step, AGMark first dynamically identifies semantic-critical evidence based on attention weights for visual relevance, together with context-aware coherence cues, resulting in a more adaptive and well-calibrated evidence-weight distribution. It then determines the proportion of semantic-critical tokens by jointly considering uncertainty awareness (token entropy) and evidence calibration (weight density), thereby enabling more reliable adaptive vocabulary partitioning to avoid irrelevant tokens.
Empirical results consistently confirm that AGMark outperforms conventional methods, substantially improving generation quality and yielding particularly strong gains in visual semantic fidelity in the later stages of generation.
Our framework maintains highly competitive detection performance (at least 99.36\% AUC) and robust attack resilience (at least 88.61\% AUC) without sacrificing inference efficiency, taking a significant step toward reliability-preserving multimodal watermarking.
\end{abstract}


\maketitle

\newcommand\kddavailabilityurl{https://doi.org/10.5281/zenodo.20339442}
\ifdefempty{\kddavailabilityurl}{}{
\begingroup\small\noindent\raggedright\textbf{Resource Availability:}\\
The source code of this paper has been made publicly available at \url{\kddavailabilityurl}.
\endgroup
}

\section{Introduction}

Large vision language models (LVLMs) demonstrate strong capabilities across a broad spectrum of downstream tasks \cite{ye2025survey, li-etal-2025-hierarchical, zhang2026llm}. 
However, their increasing deployment also introduces risks of malicious misuse, motivating the urgent need for robust intellectual property protection and reliable mechanisms to ensure content authenticity \cite{OxfordChapter,li2025construction}. 
Model watermarking addresses this by embedding identifiable signals into generated content for source attribution and traceability \cite{OxfordChapter,yi2025unified}. Existing watermarking techniques for such models can be categorized into logits-based and sampling-based methods \cite{pan2024markllm}, the former becoming the dominant line of work \cite{liu2024survey}. The pioneering method KGW \cite{kirchenbauer2023kwg} employs a secret hash key to partition the vocabulary into green and red token sets and biases the generation process toward green tokens. 
Building on this framework, follow-up studies propose detection statistics based on token entropy \cite{lee2024sweet,gu2025invisible} or the green list probability mass \cite{wang-etal-2025-morphmark}, aiming to better balance watermark detectability and generation quality.

\begin{figure}[t]
    \centering
    \includegraphics[width=\columnwidth]{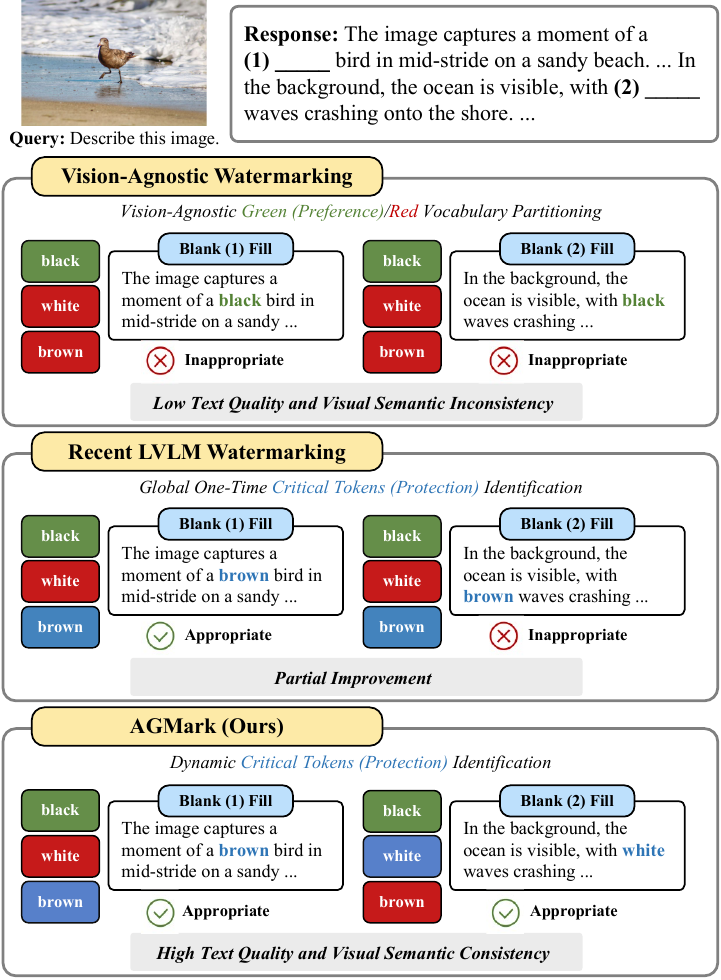}
    \caption{
    Paradigm comparison of our AGMark, vision-agnostic watermarking methods, and recent LVLM watermarking methods.
        }
    \label{fig:first}
\end{figure}
The majority of relevant watermarking techniques were developed for text-only large language models (LLMs) and do not adequately account for the challenges posed by LVLMs under multimodal conditioning \cite{pan2024markllm}. 
Both random vocabulary partitioning (logits-based paradigm) and uniform logit manipulation (sampling-based paradigm) disrupt vision-conditioned language generation by introducing vocabulary biases that contradict actual visual semantics (e.g., objects, scene descriptors). 
Hence, naively applying these methods to LVLMs often leads to perceptible degradation in the quality and fidelity of the generated text.
Some recently proposed watermarking methods~\cite{liu2025vla, zheng2026visual} tailored for LVLMs have made promising progress by embedding detectable watermarks while maintaining semantic fidelity. 
They identify the cross-modal coordination-critical weight for each vocabulary token and protect a selected portion of highly important evidence. 

However, two fundamental limitations remain: these methods compute vision-critical token weights only once at initialization, ignoring the guiding role of the density of  weight distribution when determining the proportion of protected tokens. This may cause misalignment of semantically critical evidence in later stages of generation and, under an overly concentrated weight distribution, introduce low-quality tokens from the long tail of the distribution. Figure~\ref{fig:first} illustrates these shortcomings with a concrete example, covering both vision-agnostic watermarking and recent LVLM watermarking methods, contrasted with our proposed approach.

To address these challenges, we propose Attention-Guided Dynamic Watermarking (\textbf{AGMark}), a two-stage watermarking framework that dynamically aligns visual semantics and supporting evidence at each decoding step.
First, \textbf{Semantic-Critical Weight Extraction} estimates the current semantic-critical weights by leveraging contextual coherence, and measures real-time reliance on visual semantics using attention weights over visual regions.
Second, \textbf{Adaptive Vocabulary Partitioning} increases the protected-token proportion under low uncertainty and dispersed evidence weights, thereby ensuring high-quality selection and sufficient coverage of semantic-critical tokens.
This adaptive mechanism ensures that watermark strength is concentrated on tokens strongly supported by the vision content, actively guiding the model towards visual fidelity and away from potential hallucinations, particularly under high uncertainty or highly concentrated evidence weights.

\begin{figure*}[t]
\includegraphics[width=\textwidth]{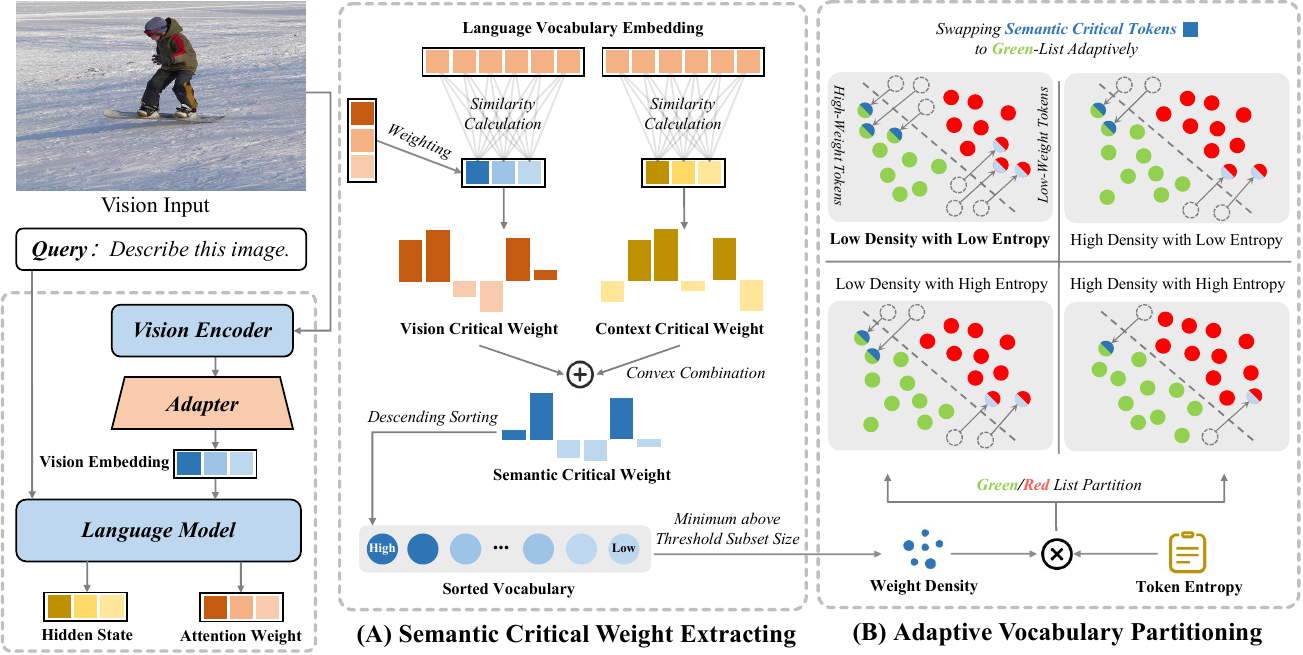}
    \caption{
        Overview of our AGMark framework, which consists of two components: (A) Semantic-Critical Weight Extraction: dynamically extracting weights via similarity calculation to fuse vision-critical weights and context-critical weights.
        (B) Adaptive Vocabulary Partitioning: leveraging token entropy and density of extracted weights to adaptively swap high evidence tokens into the green list, protecting visual fidelity.
    }
    \label{fig:method}
\end{figure*}
In summary, our contributions are as follows:
\begin{itemize}
    \item We propose a novel approach, AGMark, which provides cross-modal semantic guidance by identifying key vision-grounded semantics at each decoding step using attention weights and accounting for contextual coherence in the ongoing discourse.
    \item AGMark determines the scale of semantic-critical tokens by jointly considering uncertainty awareness and evidence calibration, enabling adaptive vocabulary partitioning and helping mitigate the preservation–detection trade-off.
    \item We conduct extensive experiments to validate AGMark in terms of text quality, visual fidelity, detectability (at least 99.36\% AUC), and robustness, with a slight and acceptable increase in inference time.
\end{itemize}

\section{Preliminaries}
\paragraph{LVLM Generation}
LVLMs commonly adopt a shared semantic mapping strategy, where visual representations are projected into the same embedding space as text tokens \cite{li-etal-2025-lost-embeddings}.
Specifically, given a model $\mathcal{M}$ with a vocabulary $\mathcal{V}$ of size $|\mathcal{V}|$, the visual input $\mathbf{v}$ is first processed by a vision encoder to obtain image features.
A multimodal adapter bridges the dimensional gap between modalities and projects the features into the shared embedding space, which aligns with the visual embedding and the textual embedding matrix.
The model subsequently generates the $t$-th token $y_t$ auto-regressively by computing a distribution over $\mathcal{V}$:
\begin{equation}
\label{equ:lvlm}
    l_t = \mathcal{M}(\mathbf{v}, y_{<t}), \quad
    p_t = \mathrm{Softmax}(l_t),
\end{equation}
where $y_{<t}$ denotes the textual context, $l_t$ is a vector of pre-softmax logits with dimensionality $|\mathcal{V}|$, and $p_t$ is the resulting next-token distribution.

\paragraph{Attention Mechanism}
The attention mechanism computes token relevance by projecting the hidden states from the previous layer into query $Q$, key $K$, and value $V$ representations using projection matrices $\mathbf{W}_Q$, $\mathbf{W}_K$, and $\mathbf{W}_V$, respectively. The attention is computed  as:
\begin{equation}
\mathrm{Attention}(Q,K,V) = \mathcal{A} \,V = \mathrm{Softmax}\!\left(\frac{Q K^\top}{\sqrt{d_K}}\right)  V,
\end{equation}
where $d_K$ denotes the key dimensionality, and we use $\mathcal{A}$ to represent the attention weight matrix.

\paragraph{Logits-based Watermarking}
Logits-based watermarking partitions the vocabulary into a red list $\mathcal{R}$ and a green list $\mathcal{G}$ according to a ratio $\gamma$ (typically $|\mathcal{G}| = \gamma|\mathcal{V}|$), which is deterministically derived by hashing the prefix $y_{<t}$ together with a secret key $\xi$. Under this paradigm, a positive bias $\delta$ is added to the logits of tokens in $\mathcal{G}$, thereby increasing their sampling probability and forming a recognizable watermark pattern.

\section{Methodology}

We propose AGMark, a vision-aligned watermarking framework that identifies semantic-critical tokens to align watermark injection with visual-grounded semantics. 
As illustrated in Figure \ref{fig:method}, our watermarking framework consists of two components: 
(A) a dynamic semantic-critical weight extraction part that fuses vision-critical weights and context-critical weights 
(Section \ref{subsec: Semantic-Critical Weight Extraction}); and
(B) a vocabulary partition that is adaptively regulated based on weight density and token entropy, which swaps high-evidence tokens into the green list (Section \ref{subsec: Adaptive Vocabulary Partitioning}).
This method achieves watermark injection based on dynamic and precise key evidence support, yielding strong detectability with improved visual fidelity.

\subsection{Semantic-Critical Weight Extraction}
\label{subsec: Semantic-Critical Weight Extraction}

Our first challenge is to effectively obtain \textit{vision-critical weights}.
We argue that the static, one-shot computation adopted by VLA-Mark \cite{liu2025vla} is suboptimal, since the visual evidence required at each generation step may vary during decoding
\cite{anderson2018bottom}.

To capture appropriate and critical visually-grounded tokens throughout auto-regressive decoding, we therefore compute token importance weights dynamically.
The model's attention weights over visual regions reflect the visual information that it focuses on, serving as a natural and informative guidance signal.

Let $\mathcal{A}_t^{\mathbf{v}}$ denote the model's attention distribution over the visual tokens at decoding step $t$, where $\mathcal{A}^{\mathbf{v}}_t(i)$ is the attention weight assigned to the $i$-th visual token. Here, $N^\mathbf{v}$ is the number of visual tokens, and $\mathcal{E}^{\mathbf{v}}$ denotes the corresponding visual token embeddings. We further denote by $\mathcal{E}^{\mathbf{t}}$ the textual embedding matrix for the entire vocabulary, where $\mathcal{E}^{\mathbf{t}}(k)$ represents the embedding of the $k$-th vocabulary token. At each decoding step $t$, the vision-critical weight $\psi_t^{\mathbf{v}}(k)$ for token $k$ is defined as:
\begin{equation}
\psi_{t}^{\mathbf{v}}(k)
=
\sum_{i=1}^{N^\mathbf{v}}
\mathcal{A}^{\mathbf{v}}_{t}(i) \,
\frac{
\mathcal{E}^\mathbf{v}(i) \, \mathcal{E}^\mathbf{t}(k)^\top
}{
\left\|\mathcal{E}^\mathbf{v}(i)\right\|
\left\|\mathcal{E}^\mathbf{t}(k)\right\|
}.
\end{equation}
This quantity corresponds to the attention-weighted aggregation of cosine similarities between visual token embeddings and the textual embedding of token $k$, thereby measuring how strongly each candidate token is aligned with the visual evidence attended to at step $t$.

However, visual saliency alone is insufficient: visual importance must be grounded in and consistent with the textual context. We use the latent representation $\mathcal{E}^\mathbf{h}$ \cite{yi2025latent} at the current decoding position (i.e., the last-layer hidden state) to compute a \textit{context-critical weight} $\psi_t^{\mathbf{c}}(k)$ for each vocabulary token $k$ as:
\begin{equation}
\psi_t^{\mathbf{c}}(k) = 
\frac{\mathcal{E}^\mathbf{h} \, {\mathcal{E}^{\mathbf{t}}{(k)}}^{\top}}
{\|\mathcal{E}^\mathbf{h}\| \, \|\mathcal{E}^{\mathbf{t}}{(k)}\|}.
\end{equation}

To harmonize importance scores from distinct sources that may follow different distributions, we apply standardization:
\begin{equation}
    z(x) = \frac{x - \mu}{\sigma + \epsilon},
\end{equation}
where $\mu$ and $\sigma$ denote the mean and standard deviation of importance scores from the corresponding source, and $\epsilon$ is a small constant added to the denominator for numerical stability.
This normalization maps scores to a common scale, facilitating subsequent aggregation while preserving the relative significance patterns within each source.


Finally, we derive the dynamic critical weights $\psi_t(k)$ by fusing the vision-critical weight $\psi_t^{\mathbf{v}}(k)$ and the context-critical weight $\psi_t^{\mathbf{c}}(k)$ via a convex combination:
\begin{equation}
    \psi_t(k) = \omega \cdot z(\psi_t^{\mathbf{v}}(k)) + (1 - \omega) \cdot z(\psi_t^{\mathbf{c}}(k)),
\end{equation}
where $\omega \in [0,1]$ controls the strength of visual importance or contextual importance. 
Then, we perform min–max normalization to each $\psi_t(k)$:
\begin{equation}
\label{equ:total norm importence}
\tilde{\psi}_t(k) = 
\frac{\psi_t(k) - \min_{i \in \mathcal{V}} \psi_t(i)}
{\max_{i \in \mathcal{V}} \psi_t(i)- \min_{i \in \mathcal{V}} \psi_t(i) +\epsilon}.
\end{equation}

We sort all tokens in the vocabulary $\mathcal{V}$ by their normalized semantic-critical weights $\tilde{\psi}_t$ in descending order, yielding a prioritized vocabulary $\mathcal{V}^*_t$ at step $t$:
\begin{equation}
\mathcal{V}^{*}_{t}
=
(i_1, i_2, \ldots, i_{|\mathcal{V}|}),
\quad
\text{where }
\tilde{\psi}_{t}(i_1)
\ge
\tilde{\psi}_{t}(i_2)
\ge
\cdots
\ge
\tilde{\psi}_{t}(i_{|\mathcal{V}|}).
\end{equation}
The tokens in the top portion of $\mathcal{V}^*_t$ are referred to as \textit{semantic-critical tokens}.

Building on the above pipeline, we assign each vocabulary token $k$ a bounded, normalized semantic-critical weight $\tilde{\psi}_t(k) \in [0,1]$, which quantifies the degree to which the current generation depends on that token. It is worth noting that this module introduces only a small number of matrix operations, resulting in a slight and acceptable increase in computational overhead. We report a detailed comparison of inference latency in Appendix \ref{apx:Efficiency}.

\subsection{Adaptive Vocabulary Partitioning}
\label{subsec: Adaptive Vocabulary Partitioning}
Next, guided jointly by \textbf{uncertainty awareness} (token entropy) and \textbf{evidence calibration} (weight density), we dynamically determine the proportion of semantic-critical tokens to be protected at the current generation step $t$, enabling adaptive vocabulary partitioning.

LVLMs infer the probability value $p_t$ of the next token based on the given multimodal input, as shown in Equation \ref{equ:lvlm}. 
Specifically, at each generation step $t$, we measure the token entropy $\mathcal{H}_t$ \cite{shannon1948mathematical}:
\begin{equation}
    \mathcal{H}_t = - \sum_{i\in\mathcal{V}} p_{t}{(i)} \log p_{t}{(i)}.
\end{equation}
The normalized entropy, which quantifies the uncertainty at each generation step, is then determined by:
\begin{equation}
\label{equ:hnorm}
    \mathcal{H}_t^{\text{norm}} = \frac{\mathcal{H}_t}{\mathcal{H}^{\text{max}}} = \frac{\mathcal{H}_t}{\log |\mathcal{V}|} \in [0,1],
\end{equation}
where $\mathcal{H}^{\text{max}}$ is the theoretical maximum value of entropy \cite{liu2025vla}. Entropy reflects the uncertainty of the model’s current generation step, where lower entropy indicates that the predicted probability mass is concentrated on a small number of candidate tokens. 

Additionally, inspired by the nucleus sampling (\textit{top-$p$}) strategy~\cite{holtzmancurioustopp}, 
we characterize the distribution density of normalized semantic-critical weights $\tilde{\psi}_t(i)$ by identifying a minimal high-importance subset of the vocabulary $\mathcal{V}$, thereby suppressing long-tail candidates that are more likely to yield low-quality or off-context tokens.
Specifically, we identify the minimal prefix subset
$\mathcal{S}_t \subseteq \mathcal{V}_t^*$ whose cumulative normalized importance mass reaches a predefined threshold $\tau$:
\begin{equation}
\mathcal{S}_t =
\left\{
\mathcal{V}_t^*(j)
\right\}_{j=1}^{m_t},
\quad
m_t =
\min \left\{
m :
\frac{
\sum_{j=1}^{m} \tilde{\psi}_t\!\left(\mathcal{V}_t^*(j)\right)
}{
\sum_{i\in \mathcal{V}} \tilde{\psi}_t(i)
}
\ge \tau
\right\}.
\end{equation}
Then, we define the weight density value $\rho_t$ as the fraction of tokens contained in this subset:
\begin{equation}
\label{equ:visualdensity}
\rho_t
=
\frac{\left|\mathcal{S}_t\right|}{|\mathcal{V}|} \in (0,1].
\end{equation}
Based on the normalized entropy $\mathcal{H}_t^{\text{norm}}$ and weight density value $\rho_t$, we calculate the semantic-critical tokens ratio $\eta_t$:
\begin{equation}
    \eta_t = \alpha \cdot\rho_t \cdot (1-\mathcal{H}_t^{\text{norm}}),
\end{equation}
where the ratio $\alpha$ controls the base semantic-critical tokens proportion. 
We derive the semantic-critical tokens set $\mathcal{C}_t \subseteq \mathcal{V}_t^*$ based on $\eta_t$:
\begin{equation}
\mathcal{C}_t =\textrm{Top}_{\left\lceil \eta_t |\mathcal{V}| \right\rceil}(\mathcal{V}_t^*) = \big( {\mathcal{V}_t^*}{(1)}, \ldots , {\mathcal{V}_t^*}{(\left\lceil \eta_t |\mathcal{V}| \right\rceil)} \big).
\end{equation}
Then, we swap $\mathcal{X}_t = \mathcal{C}_t \cap \mathcal{R}_t$ into green by removing the $|\mathcal{X}_t|$ lowest semantic-critical weight tokens $\mathcal{Y}_t \subseteq \mathcal{G}_t$:
\begin{equation}
\begin{split}
\mathcal{G}_t &\leftarrow (\mathcal{G}_t \setminus \mathcal{Y}_t)\cup \mathcal{X}_t,\\
\mathcal{R}_t &\leftarrow (\mathcal{R}_t \setminus \mathcal{X}_t)\cup \mathcal{Y}_t,
\end{split}
\end{equation}
optionally gating the swap by a margin threshold and a per-step cap to avoid oscillation.

The watermarked probability distribution is computed following the logits-based paradigm \cite{kirchenbauer2023kwg}, which adds a positive bias $\delta$ to the logits of tokens in $\mathcal{G}$, thereby increasing their sampling probability:
\begin{equation}
\hat{p}_t{(k)}=
\begin{dcases}
\frac{\exp\left(l_t{(k)}+\delta\right)}
{\sum_{i\in \mathcal{R}_t}\exp\left(l_t{(i)}\right)+\sum_{i\in \mathcal{G}_t}\exp\left(l_t{(i)}+\delta\right)}, 
& k\in \mathcal{G}_t,\\
\frac{\exp\left(l_t{(k)}\right)}
{\sum_{i\in \mathcal{R}_t}\exp\left(l_t{(i)}\right)+\sum_{i\in \mathcal{G}_t}\exp\left(l_t{(i)}+\delta\right)}, 
& k\in \mathcal{R}_t.
\end{dcases}
\end{equation}
Here, $\hat{p}_t{(k)}$ denotes the watermarked probability of sampling token $k$ at step $t$.

We expand the coverage of semantic-critical tokens in low-entropy and dispersed weight distribution scenarios, thereby preserving the fidelity of the generated text. 
In other cases, the coverage of semantic-critical tokens is adaptively reduced to varying degrees to enhance robustness. This dual-signal design aims to balance the trade-off between watermark embedding strength and generation quality.

\begin{table*}[th]
\caption{Performance comparison in terms of metrics AUC $\uparrow$, Accuracy (ACC) $\uparrow$, Perplexity (PPL) $\downarrow$, BLEU $\uparrow$, BertScore (BertS) $\uparrow$, and STS $\uparrow$. 
\textbf{Bold} values indicate the best performance among all methods, while \underline{underlined} values indicate the second best.
}
\centering
\newcolumntype{C}{>{\centering\arraybackslash}X}
\begin{tabularx}{\textwidth}{clCCCCCCCCCCCC}
\toprule
\multirow{2}[2]*{\textbf{Model}} & \multirow{2}[2]*{\textbf{Method}} & \multicolumn{6}{c}{\textbf{AMBER}} & \multicolumn{6}{c}{\textbf{MS-COCO}} \\
\cmidrule(lr){3-8} \cmidrule(lr){9-14} 
 & & AUC & ACC & PPL & BLEU & BertS & STS & AUC & ACC & PPL & BLEU & BertS & STS \\
\midrule
\multirow{6}{*}{\rotatebox{90}{\textbf{Llava-Next}}}  
&KGW & \textbf{99.99}&\textbf{99.75}&5.29&22.88&90.00&89.43 &\textbf{99.93}&\underline{98.50}&5.81&24.72&89.35&87.75   \\
&SynthID & \underline{99.91}&98.50&5.29&25.93&90.28&89.79  & 99.88&\underline{98.50}&5.33&\underline{26.37}&\underline{90.28}&\underline{89.87}   \\
&IE & 99.58&97.38&5.56&22.37&90.11&89.64 & 99.78&\textbf{98.88}&5.58&20.52&89.74&88.39  \\
&MorphMark &99.67&97.38&\underline{4.98}&\underline{26.92}&\underline{90.44}&90.07 & 99.75&97.75&\underline{5.07}&25.62&90.24&89.53   \\
&VLA-Mark & 99.13&96.38 & 5.07&25.98&90.34&\underline{90.12}& 99.68&98.13&5.22&20.66&89.53&87.63   \\
&AGMark (Ours) & 
\underline{99.91}&\underline{99.50}&\textbf{4.80}&\textbf{31.88}&\textbf{91.14}&\textbf{91.85}
&\underline{99.91}&\textbf{98.88}&\textbf{4.95}&\textbf{29.86}&\textbf{90.76}&\textbf{90.51} \\
\midrule
\multirow{6}{*}{\rotatebox{90}{\textbf{Qwen3-VL}}}  
&KGW & \textbf{99.46}&\textbf{96.25}&\underline{5.83}&27.66&90.74&92.54 & 99.74&\underline{98.00}&5.88&27.28&90.73&92.34  \\
&SynthID&
99.12&95.88&6.32&26.60&90.71&92.33&
99.77&\underline{98.00}&6.70&24.80&90.41&91.42 \\
&IE&99.11&96.00&6.03&27.98&90.93&92.32 & \underline{99.78}&\textbf{98.25}&6.26&24.87&90.39&91.65 \\
&MorphMark & 99.27&95.75&\textbf{5.81}&28.87&90.93&92.93  & 
99.64&97.25&\underline{5.87}&28.21&90.81&92.43  \\
&VLA-Mark &99.05&96.00&5.96&\underline{31.12}&\underline{91.40}&\underline{93.72} & 
99.69&97.32&6.04&\underline{31.10}&\underline{91.22}&\underline{92.91}   \\
&AGMark (Ours) &\underline{99.36}&\underline{96.22}&\underline{5.83}&\textbf{31.88}&\textbf{91.46}&\textbf{93.99} & \textbf{99.82}&97.88&\textbf{5.85}&\textbf{31.95}&\textbf{91.45}&\textbf{93.81}  \\
\midrule
\multirow{6}{*}{\rotatebox{90}{\textbf{InternVL-3.5}}}  
&KGW &\underline{99.57}&\textbf{97.63}&4.64&25.99&90.16&90.67 & \underline{99.65}&\underline{97.63}&4.62&27.78&90.25&\underline{90.59}   \\
&SynthID& \textbf{99.69}&96.96&4.65&\underline{27.27}&90.21&90.89 &99.49&97.25&4.60&\underline{28.38}&90.34&90.11  \\
&IE & 99.19&97.00&\underline{4.52}&20.82&88.75&87.12 & 99.59&97.38&\underline{4.54}&23.50&89.48&88.33  \\
&MorphMark & 99.56&97.38&4.56&27.20&\underline{90.31}&90.81 & \textbf{99.68}&\underline{97.63}&4.56&27.57&90.32&90.56  \\
&VLA-Mark & 99.23&97.00&4.72&25.00&90.17&\underline{91.03} &99.58&96.75&4.55&27.86&\underline{90.42}&90.44    \\
&AGMark (Ours) & \underline{99.57}&\underline{97.50}&\textbf{4.50}&\textbf{27.45}&\textbf{90.40}&\textbf{91.12} &
\underline{99.65}&\textbf{97.85}&\textbf{4.53}&\textbf{30.88}&\textbf{91.08}&\textbf{91.74}   \\
\bottomrule
\end{tabularx}
\label{tab:main}
\end{table*}

\section{Experimental Setup}

\paragraph{Models and Datasets} We assess our method on three mainstream 8B LVLMs, including Llava-Next-Llama3~\cite{liu2024llavanext}, Qwen3-VL~\cite{qwen3technicalreport}, and InternVL-3.5~\cite{wang2025internvl3}. Performance is evaluated using the AMBER~\cite{wang2023amber} and MS-COCO~\cite{lin2014microsoftcoco} datasets, following prior studies~\cite{liu2025vla, zheng2026visual}. Unless otherwise specified, Llava-Next-Llama3 and AMBER are used as the default model and dataset, respectively.

\paragraph{Evaluation Metrics} 
Our evaluation spans detectability performance (AUC and Accuracy), visual consistency (CHAIR \cite{wang2023amber}), text quality, and robustness against five types of attack \cite{lau-etal-2024-waterfall, liu2025vla}, which alter text through word insertion, deletion, synonym substitution, paraphrasing, and translation. Text quality is assessed in terms of linguistic quality (Perplexity \cite{shannon1948mathematical} and BLEU \cite{papineni-etal-2002-bleu}) and semantic alignment (BertScore \cite{zhangbertscore} and STS \cite{cer-etal-2017-sts}). 

\paragraph{Baselines} We compare our approach against five representative model watermarking baselines, including KGW \cite{kirchenbauer2023kwg}, SynthID \cite{dathathri2024scalable}, IE \cite{gu2025invisible}, MorphMark \cite{wang-etal-2025-morphmark}, and VLA-Mark \cite{liu2025vla}. These methods are selected for their strong emphasis on detection performance and text quality. Notably, VLA-Mark is specifically designed for LVLMs. Most implementations are based on the MarkLLM \cite{pan2024markllm} repository. 

More relevant details are reported in Appendix \ref{sec:Experimental Details}. 

\section{Results and Analysis}

\subsection{Main Results}
\begin{figure*}[ht]
    \centering
    \includegraphics[width=\textwidth]{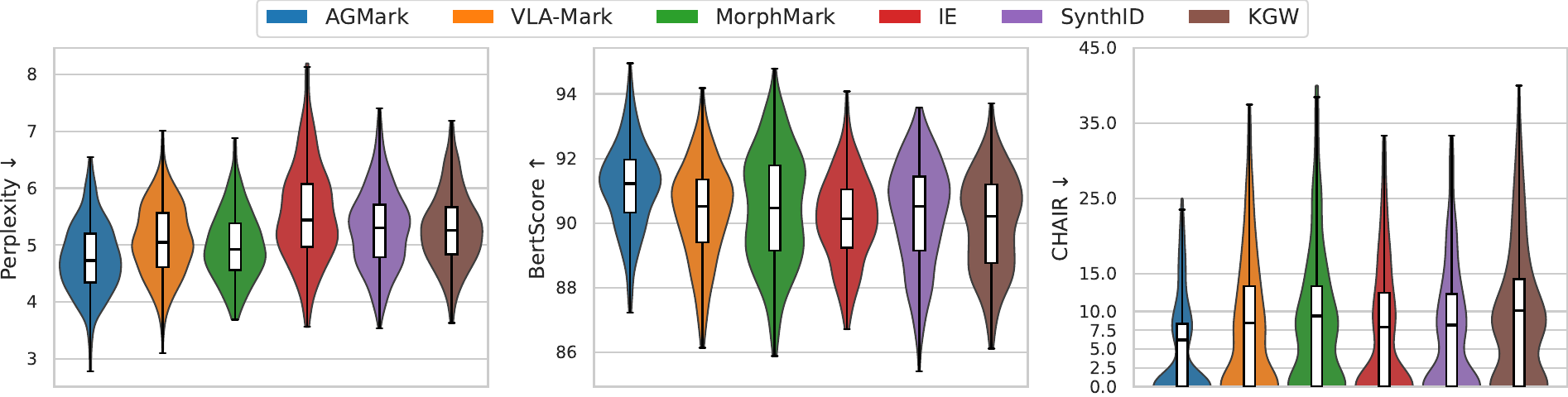}
    \caption{
        Violin plots showing the sample-wise performance distribution including Perplexity (Left), BertScore (Middle), and CHAIR (Right). 
        For Perplexity and BertScore, the overlaid box plots indicate the median. For CHAIR, however, the box plot shows the mean, since the large number of zero values for AGMark and IE would otherwise yield a median of 0.
    }
    \label{fig:distribution_3metrics}
\end{figure*}

Table \ref{tab:main} provides a detailed performance comparison of AGMark with several baseline methods across three LVLMs. 
The results empirically validate our primary hypothesis: while vision-agnostic watermarking mechanisms degrade generation quality, our dynamically identified semantic-critical tokens actively preserve and enhance it while maintaining strong detectability.

\begin{figure*}[th]
    \centering
    \includegraphics[width=\textwidth]{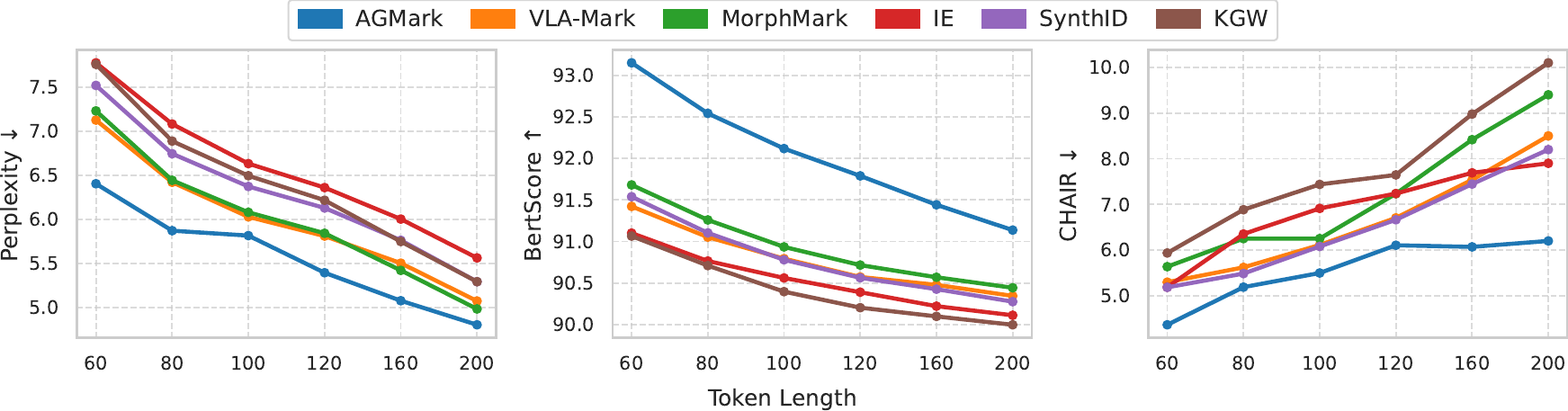}
    \caption{
        Line charts depict length-wise performance retention across response token lengths ranging from 60 to 200 tokens, evaluated using three metrics: Perplexity (Left), BertScore (Middle), and CHAIR (Right).
    }
    \label{fig:maintenance_3metrics}
\end{figure*}
\textbf{In terms of text quality}, AGMark achieves the best performance in the vast majority of settings.
For instance, on the Llava-Next backbone evaluated using the AMBER dataset, it significantly outperforms the second-best method, MorphMark, in linguistic quality. Specifically, AGMark reduces perplexity by 0.18 (4.80 vs. 4.98) and improves BLEU by 4.96 percentage points (31.88\% vs. 26.92\%). Notably, AGMark is the only method that exceeds 91\% on both semantic alignment metrics.
Crucially, these improvements do not come at the cost of detectability.

\textbf{Regarding detectability}, AGMark achieves an AUC of at least 99.36\% and an accuracy of at least 96.22\% across three models and two datasets. Notably, AGMark consistently ranks among the top two methods in terms of AUC across the vast majority of settings, demonstrating strong detection reliability.
This confirms that accounting for visual evidence in the watermarking process effectively aligns the generated text with visual content without compromising the watermark’s statistical detectability.

\subsection{Sample-wise Performance Distribution}
We report sample-level performance distributions in Figure \ref{fig:distribution_3metrics} to examine variability and long-tail behaviors that are not captured by aggregate metrics. This analysis further highlights stability across individual samples, especially in challenging cases where performance degradation may be more pronounced.

In the left panel, AGMark shows a lower median perplexity than the other baselines, with the lowest minimum and maximum. In the middle panel, AGMark achieves a higher median BertScore. MorphMark attains a slightly higher maximum, but also exhibits a much lower minimum. These results suggest that AGMark better preserves the original model’s language distribution, improving linguistic quality and semantic alignment.
In the right panel, AGMark, IE, and SynthID are the top three methods with the lowest hallucination score (CHAIR). AGMark achieves the lowest mean (6.2\% vs. 7.9\% and 8.2\%) and also the lowest maximum value, indicating better performance in both average and worst-case settings. This strong hallucination reduction suggests that AGMark effectively improves visual grounding and alignment. 

Overall, this demonstrates the broad applicability of our method: by dynamically and accurately identifying semantically critical tokens, AGMark not only achieves strong average performance but also avoids severe performance degradation when handling challenging samples.

\subsection{Length-wise Performance Maintenance}

We further analyze the impact of watermarking on text quality and visual fidelity across varying generation lengths, ranging from 60 to 200 tokens, as shown in Figure~\ref{fig:maintenance_3metrics}. 
This setting allows us to examine whether different watermarking methods remain stable as generation proceeds, especially when longer outputs impose greater challenges on text quality and visual consistency.

As the generation length increases, all methods exhibit rising CHAIR scores accompanied by a decline in BertScore, suggesting that longer generation horizons exacerbate visual inconsistency and semantic drift under watermarking. 
Although perplexity decreases in later stages, this trend mainly reflects improved linguistic quality and does not indicate better semantic alignment or visual consistency.
Beyond their generally inferior performance, some baseline methods show pronounced sensitivity to generation length, revealing weaknesses in robustness. 
For instance, IE attains the second-best visual--semantic alignment at 200 tokens, yet drops to the second-worst rank in the 80--100 token range.

In contrast, AGMark consistently maintains superior text quality and visual consistency. 
Notably, in the range of 120--200 tokens, AGMark demonstrates an exceptionally low rate of hallucination increase. 
By dynamically allocating and preserving semantically critical tokens, AGMark minimizes distortions in fluency and coherence, thereby enabling more accurate generation in the later stages.

\begin{table*}[ht]
\caption{Ablation study comparing the full AGMark algorithm to its variants lacking specific components. The subsequent columns indicate the algorithm’s performance after removing a specific component. }
\centering
\newcolumntype{C}{>{\centering\arraybackslash}X}
\begin{tabularx}{\textwidth}{lCCCCCCC}
\toprule
\multirow{2}[2]*{\textbf{Ablation}} & \multirow{2}[2]*{\textbf{Full}} & 
\multicolumn{3}{c}{\textbf{Weight Extraction}} 
& \multicolumn{3}{c}{\textbf{Vocabulary Partitioning}} 
\\
\cmidrule(lr){3-5}  \cmidrule(lr){6-8}
& & w/o Attention & w/o Vision &  w/o Context & w/o Entropy &  w/o Density &  Fixed-Scale \\
\midrule
CHAIR $\downarrow$& 6.20 & $8.50_{+2.30}$ & $7.90_{+1.70}$ & $7.40_{+1.20}$ & $7.60_{+1.40}$ & $8.10_{+1.90}$ &  $8.40_{+2.20}$ \\
Perplexity $\downarrow$& 4.80 & $5.17_{+0.37}$ & $5.68_{+0.88}$ & $5.80_{+1.00}$ & $4.92_{+0.12}$ & $4.86_{+0.06}$ & $5.18_{+0.38}$\\
BLEU &31.88 & $29.63_{-2.25}$ & $18.25_{-13.63}$ & $17.94_{-13.94}$&  $28.11_{-3.77}$  & $28.65_{-3.23}$ & $27.63_{-4.25}$ \\
BertScore & 91.14 & $90.75_{-0.39}$& $89.44_{-1.70}$ & $89.36_{-1.78}$ & $90.66_{-0.48}$ & $90.73_{-0.41}$ & $90.41_{-0.73}$\\
STS &91.85&$90.47_{-1.38}$ &$88.20_{-3.65}$&$88.05_{-3.80}$& $90.50_{-1.35}$& $90.63_{-1.22}$ & $89.99_{-1.86}$ \\
\bottomrule
\end{tabularx}
\label{tab:ablation}
\end{table*}

\subsection{Ablation Study}
We validate the architectural effectiveness of our framework: 
The strategy for semantic-critical weight extraction (Section \ref{subsec: Semantic-Critical Weight Extraction}) and the structural necessity of our adaptive components (Section \ref{subsec: Adaptive Vocabulary Partitioning}). 
We construct several variants of AGMark by removing these components.
For the weight extraction component, we completely remove the attention guiding to obtain \textit{w/o Attention}, the vision-critical weight to obtain \textit{w/o Vision} ($\omega = 0.00$), and the context-critical weight to obtain \textit{w/o Context} ($\omega = 1.00$). 
For the adaptive vocabulary partitioning component, we completely remove token entropy or weight density to obtain \textit{w/o Entropy} and \textit{w/o Density}. 
In addition, we obtain \textit{Fixed-Scale} by removing the adaptive vocabulary partitioning module and using a fixed scale for semantic-critical tokens.

The results in Table \ref{tab:ablation} validate the critical role of individual components in AGMark’s design and reveal some significant insights: 
\textbf{(1) Necessity of dynamic and dual-aspect weight extraction}: Removing any component leads to a consistent degradation across all metrics. In particular, ablating attention guiding causes the most severe loss in vision–language semantic fidelity (+2.30 CHAIR), highlighting the importance of dynamically providing appropriate visual evidence at each generation step. The visual critical weighting anchors the output to visual semantics, but it can overlook narrative structure and relational plausibility. In contrast, the context-critical weighting helps maintain a coherent narrative trajectory and logical consistency, compensating for this limitation. For example, it recognizes that “waves crashing onto the shore” is correct, whereas “waves crashing onto the waves” is nonsensical, even if “waves” is visually salient evidence. 
\textbf{(2) Adaptive vocabulary partitioning contribution}: Ablating uncertainty awareness primarily degrades textual quality. Removing evidence calibration substantially disrupts vision--language semantic alignment, and removing both leads to an even more pronounced performance drop. These results further confirm that applying a fixed protected-token budget without accounting for model uncertainty and the distribution of semantic-critical weights can introduce erroneous markings under high-entropy states. In contrast, our adaptive mechanism effectively mitigates this risk.

\subsection{Hyperparameter Analysis}

\begin{table}[ht]
\caption{The hyperparameter analysis for Llava-Next on the AMBER dataset. The results demonstrate a trade-off between detectability and visual fidelity. \textbf{Bold} values indicate the best performance among all settings. 
}
\centering
\newcolumntype{C}{>{\centering\arraybackslash}X}
\begin{tabularx}{\columnwidth}{lCCCC}
\toprule
$\omega$ &0.30 &  0.40 & 0.50* & 0.60\\
\midrule
AUC & \textbf{100.00} & 99.99  & 99.91 & 99.94\\
Accuracy & \textbf{99.88} &99.38 & 99.50 & 99.25\\
CHAIR $\downarrow$ & 6.70  & \textbf{5.60} & 6.20 & 6.80\\
Perplexity $\downarrow$  & 5.17 & 4.87& \textbf{4.80} & 4.82\\
BertScore & 90.32 & 90.67 & 91.14 & \textbf{91.35}\\
\midrule
\midrule
$\alpha$ & 0.23 & 0.25 & 0.27* & 0.29 \\
\midrule
AUC  & \textbf{99.97} &99.94&99.91 & 99.85  \\
Accuracy &99.38 &\textbf{ 99.50}  & \textbf{99.50} &  98.88  \\
CHAIR $\downarrow$& 7.20 & 7.20 & 6.20 & \textbf{5.90} \\
Perplexity $\downarrow$& 4.82 & 4.79 & 4.80 & \textbf{4.78 }\\
BertScore & 90.95 & 91.03 & 91.14 & \textbf{91.27}\\
\midrule
\midrule
$\tau$ & 0.90 & 0.95 & 0.98* & 0.99 \\
\midrule
AUC & \textbf{99.98} & 99.95 & 99.91 & 99.87\\
Accuracy & \textbf{99.50} & 99.38 & \textbf{99.50} & 99.00 \\
CHAIR $\downarrow$ & 7.60 & 7.70 & 6.20 &\textbf{5.80}\\
Perplexity $\downarrow$&  4.80 & \textbf{4.79} & 4.80 & 4.82 \\
BertScore & 90.83 & 91.04 & 91.14 & \textbf{91.22} \\
\bottomrule
\end{tabularx}
\label{tab:hyperparameter}
\end{table}
\begin{figure*}[th]
    \centering
    \includegraphics[width=\textwidth]{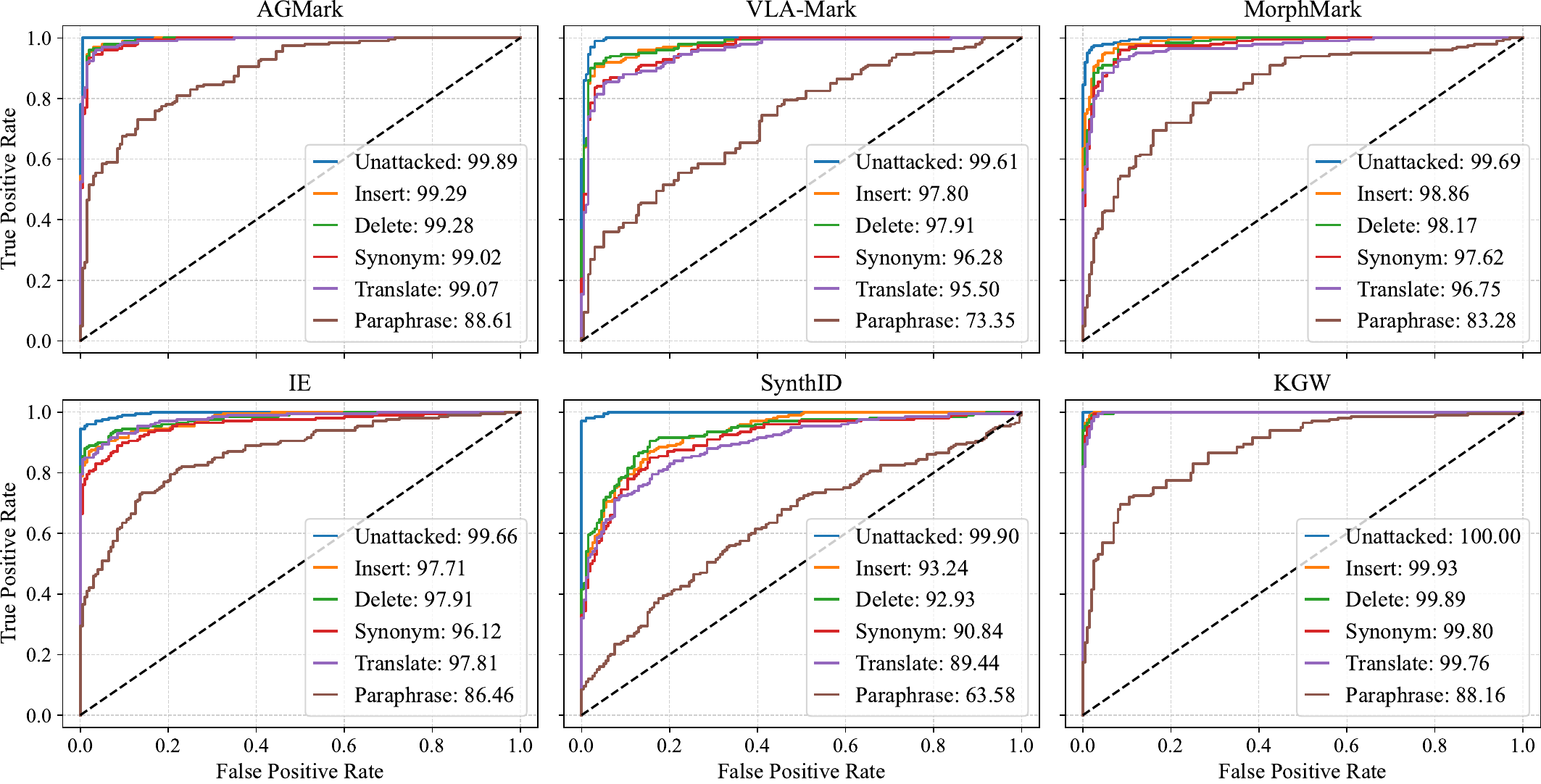}
    \caption{
        ROC curves and AUC values for six watermarking methods under various attack scenarios on the first 200 samples (hence possible slight differences from the full-set results reported in Table~\ref{tab:main}). AUC values are reported in the legends. Smaller differences between pre-attack and post-attack AUC values indicate stronger robustness against the corresponding attacks. 
        A comparison of relative performance drops is given in Appendix~\ref{apx:Relative Performance Drop}.
    }
    \label{fig:attacks}
\end{figure*}
As shown in Table~\ref{tab:hyperparameter}, we investigate the impact of three critical hyperparameters in AGMark, including the fusion-strength controller $\omega$, the base proportion of semantic-critical tokens $\alpha$, and the predefined threshold $\tau$. 
Hyperparameters $\alpha$ and $\tau$ present a clear trade-off between detectability (AUC and Accuracy) and generation quality (CHAIR, Perplexity, and BertScore), whereas the effect of $\omega$ is non-monotonic.

Larger values of $\alpha$ and $\tau$ generally tend to improve generation quality and reduce hallucinations.
For example, when $\alpha = 0.29$, Perplexity and CHAIR decrease to 4.78 and 5.90\%, respectively. This trend supports our design rationale. Larger $\alpha$ and $\tau$ allow more semantic-critical tokens to bypass random red-list exclusion, thereby protecting key content from watermark-induced perturbations and preserving generation quality. However, this protection reduces the effective space available for watermark signaling, which can weaken detectability.

In contrast, the influence of $\omega$ is non-monotonic, suggesting that careful tuning is necessary to balance visual guidance and contextual appropriateness. When $\omega$ is too small or too large, the importance weighting becomes dominated by a single modality and performance degrades. 
For instance, at $\omega = 0.30$ and $\omega = 0.60$, Perplexity increases to 5.17 and 4.82, indicating reduced fluency. 
Additionally, relying solely on visual similarity can misidentify visual neighbors as key entities, while relying only on contextual signals can induce grounding drift in long sequences~\cite{li2025mitigating}. Consistent with this, hallucinations increase at extreme settings. CHAIR rises to 6.70\% when $\omega = 0.30$, and it reaches 6.80\% when $\omega = 0.60$. These results indicate that visual relevance and contextual consistency should be jointly considered in the decision process, which aligns with prior work~\cite{chen2025mitigatingdlc}.

However, the results also highlight that excessive values for either parameter compromise detection performance. Over-prioritizing semantic tokens or over-prioritizing visually salient cues disrupts the statistical randomness required for the watermark detector, and leads to a decline in detectability efficiency (e.g., Accuracy drops to 98.88\% and 99.25\% when $\alpha$ = 0.29 and $\omega = 0.60$, respectively). 
Consequently, we identify the configuration of $\omega = 0.50$, $\alpha$ = 0.27, and $\tau$ = 0.98 as the optimal equilibrium. 

\subsection{Robustness against Attacks}

To assess the resilience of our watermark, we evaluate AGMark against five standard text-space attacks: random word insertion, deletion, synonym substitution, paraphrasing, and translation. 
Implementation details are provided in Appendix \ref{sec:Experimental Details}.

Figure \ref{fig:attacks} presents the ROC curves and AUC metrics, illustrating 
AGMark’s superior resilience, maintaining high detectability under all attacks. 
Specifically, AGMark achieves an AUC of 88.61\% under paraphrasing attacks, which yields the largest AUC drop of 11.28 percentage points among all attack settings. For the other attack types, the AUC drop is at most 0.87 percentage points under synonym attacks. In contrast, VLA-Mark, MorphMark, and IE suffer AUC drops of 3.34, 2.08, and 3.55 percentage points under synonym attacks, and drops of 26.36, 16.46, and 13.25 percentage points under paraphrasing attacks, respectively.
Moreover, AGMark outperforms KGW under paraphrasing attacks and remains competitive under the other attack types, demonstrating strong robustness to watermark removal. We attribute these improvements to dynamic extraction of semantically salient cues and adaptive vocabulary partitioning, which keep the detector anchored to key evidence and preserve detectability even under substantial lexical and structural shifts.

\section{Related Work}
\subsection{Watermarking}
Mainstream model watermarking methods predominantly focus on unimodal text generation \cite{liu2025vla}, and can be broadly categorized into two paradigms: logits-based and sampling-based \cite{wang-etal-2025-trade, yi2025latent}.

\textit{Logits-based} watermarking methods modify the token probability distribution over the vocabulary during decoding. 
The pioneering KGW \cite{kirchenbauer2023kwg} method uses a hash key to divide the vocabulary into red and green lists, favoring green tokens in the output. IE \cite{gu2025invisible} trains an entropy navigator that reduces the injection intensity as the number of green tokens increases under low entropy conditions. MorphMark \cite{wang-etal-2025-morphmark} demonstrates that the sum of probabilities
of green tokens in the watermark plays a crucial role in balancing the trade-off between injection intensity and text quality, determining the injection intensity based on the probability of green tokens.
Meanwhile, \textit{sampling-based} watermarking methods alter the sampling process by leveraging pseudorandom sequences to guide token selection. 
\citet{christ2024undetectable} propose a binary sampling method for LLMs based on pseudo-random number generation, though it is unsuitable for real-world LLMs. SynthID \cite{dathathri2024scalable} introduces a tournament-based strategy for token generation. In contrast, current research in sampling-based watermarking is limited, indicating room for advancement \cite{liu2024survey}.

Additionally, two recent logits-based watermarking methods for LVLMs, VLA-Mark \cite{liu2025vla} and VISA-Mark \cite{zheng2026visual}, both follow a common design: they compute visual-importance weights and balance watermark strength via an entropy-sensitive mechanism to guide watermark injection. However, their static, global, one-shot importance estimation limits the ability to identify semantic-critical tokens in a more fine-grained and accurate manner. 
Our proposed AGMark leverages attention-guided visual relevance to identify the most pertinent visual evidence and incorporates contextual coherence to dynamically and precisely protect semantic-critical tokens at each decoding step, thereby achieving a balanced trade-off between watermark detectability and generation quality.

\subsection{Watermark Removal Attacks}
Watermarks face the threat of being removed by malicious attackers in real-world applications \cite{OxfordChapter}. Attacks against watermarking algorithms can be broadly categorized into text manipulations and informed watermark attacks.

\textit{Text manipulations} involve traditional NLP techniques for straightforward text manipulation, such as word deletion, substitution, or insertion \cite{lau-etal-2024-waterfall}. Translation is also a common form of attack, and watermarks lack consistency in their detectability across different languages \cite{he2024can}. 
By altering the word-level structure of the text, these methods attempt to distort or eliminate the watermark.
In contrast, document-level attacks introduce more substantial changes to content and discourse structure, with paraphrasing being the most prevalent approach \cite{liu2024survey}.
Paraphrasing attacks can be carried out either manually \cite{kirchenbauer2023kwg} or using large language models \cite{chengrevealing2025}. More broadly, paraphrasing includes both monolingual rewriting and its cross-lingual variant realized through translation (optionally followed by back-translation), which has been shown to effectively degrade watermark detectability \cite{lau-etal-2024-waterfall, liu2024survey}.

A representative example of \textit{informed watermark attacks} is watermark stealing \cite{wu2024bypassing, jovanovic2024watermark}, where an attacker queries the watermarking algorithm to reverse-engineer or approximate the watermarking rule, estimate the watermark distribution of the target model, and infer whether each token conforms to the watermark constraint via a large number of carefully designed prefix queries conditioned on the context. 
However, such methods typically assume essentially unlimited access to the watermarked model, which limits their practicality in real-world settings. 


\section{Conclusion}

In this paper, we present AGMark, a vision-language aligned dynamic watermarking framework that harmonizes intellectual property protection with cross-modal semantic fidelity. By combining attention-guided extraction of dynamic semantic-critical weights with an adaptive vocabulary partitioning strategy that incorporates uncertainty awareness and evidence calibration, our method addresses the long-standing trade-off between watermark detectability and visual–semantic consistency in existing approaches. Empirical results demonstrate that AGMark achieves clear advantages, delivering strong detectability and high robustness while also improving overall visual fidelity and text quality. This work establishes a novel paradigm for watermarking in LVLMs, and we hope AGMark will help promote the responsible use of artificial intelligence technologies.

\begin{acks}
This work was supported by the New Generation Artificial Intelligence National Science and Technology Major Project (Grant No. 2025ZD0123402), the Computational Biology Program (Grant No. 25JS2830402) of Science and Technology Commission of Shanghai Municipality (STCSM), and Shanghai Municipal Science and Technology Major Project (Grant No. 2025SHZDZX025G06). 
\end{acks}
\bibliographystyle{ACM-Reference-Format}
\balance
\bibliography{main}
\appendix

\begin{table*}[ht]
\caption{Inference latency (average generation time cost in seconds) for different watermarking methods across LVLMs.}
\centering
\newcolumntype{C}{>{\centering\arraybackslash}X}
\begin{tabularx}{\textwidth}{lcCCCCCc}
\toprule
Model & AGMark (Ours) & VLA-Mark & MorphMark & IE & SynthID & KGW & unwatermarked  \\
\midrule
Llava-Next & 13.6288 & 13.5321 & 12.9452 & 15.0930 & 13.2575  & 12.8765& 12.6046 \\
InternVL-3.5 & 18.2620 & 18.1217 & 17.4910 & 19.6789 & 18.7429 & 17.3230 & 17.1905 \\
Qwen3-VL& 21.8015 &  21.5862 & 21.6214 & 22.9254 & 22.7037 & 21.5226   & 21.3882 \\
\bottomrule
\end{tabularx}
\label{tab:Efficiency}
\end{table*}

\begin{table*}[ht]
\caption{AUC drop from the unattacked baseline under adversarial attacks. \textbf{Bold} values indicate the best performance among all methods, while \underline{underlined} values indicate the second best.}
\centering
\newcolumntype{C}{>{\centering\arraybackslash}X}
\begin{tabularx}{\textwidth}{lcCCCCC}
\toprule
Method & Unattacked & Insert & Delete & Synonym & Translate & Paraphrase \\
\midrule
KGW & 0.00 (100.00) & \textbf{0.07 (99.93)} & \textbf{0.11 (99.89)} & \textbf{0.20 (99.80)} & \textbf{0.24 (99.76)} & \underline{11.84 (88.16)} \\
SynthID & 0.00 (99.90) & 6.66 (93.24) & 6.97 (92.93) & 9.06 (90.84) & 10.46 (89.44) & 36.32 (63.58) \\
IE & 0.00 (99.66) & 1.96 (97.71) & 1.76 (97.91) & 3.55 (96.12) & 1.86 (97.81) & 13.25 (86.46) \\
MorphMark & 0.00 (99.69) & 0.83 (98.86) & 1.52 (98.17) & 2.08 (97.62) & 2.95 (96.75) & 16.46 (83.28) \\
VLA-Mark & 0.00 (99.61) & 1.82 (97.80) & 1.71 (97.91) & 3.34 (96.28) & 4.13 (95.50) & 26.36 (73.35) \\
AGMark (Ours)& 0.00 (99.89) & \underline{0.60 (99.29)} &  \underline{0.61 (99.28)} & \underline{0.87 (99.02)}& \underline{0.82 (99.07)} & \textbf{11.28 (88.61)}\\
\bottomrule
\end{tabularx}
\label{tab:Drop}
\end{table*}

\section{Experimental Details}
\label{sec:Experimental Details}
\paragraph{Environments}
Experiments are conducted on a system equipped with six NVIDIA GeForce RTX 3090 GPUs with 24 GB of memory each, and two Quadro RTX 8000 GPUs with 48 GB of memory each.

\paragraph{Hyperparameter Settings}
For a fair comparison, we standardize the hyperparameters across methods:
\begin{itemize}
    \item For the main hyperparameters, we follow the default settings of the MarkLLM~\cite{pan2024markllm} repository.
    \item We set $\gamma = 0.5$ to keep the size of the green vocabulary consistent across different watermarking methods.
    \item To ensure a controlled comparison and avoid imbalanced watermark strength, we tune the watermark injection strength (i.e., $\delta$) to achieve an AUC of at least 99\%.
\end{itemize}

\paragraph{Detector}
Our detector framework follows the architectural design of VLA-Mark~\cite{liu2025vla}. 
For watermark verification, we reconstruct the generation-time logits, token entropy, and visual attention weights at each token position using the original image–text pair. 
We adopt the evaluation protocol established by VLA-Mark, using scikit-learn~\cite{pedregosa2011scikit} to compute AUC and MarkLLM's~\cite{pan2024markllm} built-in accuracy evaluator with the best decision rule to assess accuracy.

\paragraph{Seed}
We use seed 42 as the default seed for all experiments.

\paragraph{Watermarking Implementation Details}
We sample 400 examples from each of AMBER~\cite{wang2023amber} and MS-COCO~\cite{lin2014microsoftcoco}.
Following VLA-Mark \cite{liu2025vla}, we generate responses via greedy decoding and cap the output length at 200 tokens, and we perform watermark detection using the same procedure.

\paragraph{Attack Implementation Details}
The watermark removal attacks are applied to the first 200 generated samples. Following standard protocols \cite{zheng2026visual, liu2025vla, lau-etal-2024-waterfall}, we perform 25\% token-level edits on the generated responses, conduct round-trip translation on the watermarked text (English $\rightarrow$ Spanish $\rightarrow$ English), and paraphrase the watermarked text using Llama-3.1-8B-Instruct \cite{dubey2024llama3}.

\paragraph{Ablation Study Implementation Details} 
For the \textit{Fixed-Scale} variant, we apply fixed values derived from the average size under standard conditions after rounding.

\section{Efficiency}
\label{apx:Efficiency}

Table \ref{tab:Efficiency} shows the end-to-end generation latency for 10 images and 200 tokens on three LVLMs. Our AGMark adds only a small overhead over existing text-only watermarking methods. While AGMark introduces a moderate latency increase compared to lightweight baselines like KGW, the additional overhead is manageable (e.g., approx.\ +0.41s on Qwen3-VL relative to the unwatermarked baseline). This trade-off is justified by the significant gains in generation quality.

\section{Relative Performance Drop under Attacks}
\label{apx:Relative Performance Drop}

The watermark removal attacks are applied to the first 200 samples. Table \ref{tab:Drop} quantifies AGMark’s resilience through relative AUC drops under five attack scenarios.

\section{Unwatermarked Performance}

We include the unwatermarked LVLMs for additional reference with regard to the main quality metrics. 
On the AMBER dataset, BLEU, BertScore, and STS are computed using the original unwatermarked text as the reference, and therefore already capture the quality gap between watermarked and unwatermarked generations. 
    We report the Perplexity and CHAIR scores of the unwatermarked LVLMs: Llava-Next obtains 4.66 Perplexity and 5.90\% CHAIR, Qwen3-VL obtains 5.50 Perplexity and 7.10\% CHAIR, and InternVL-3.5 obtains 4.07 Perplexity and 6.70\% CHAIR.

\end{document}